\documentclass[10pt,twocolumn,letterpaper]{article}

\usepackage[pagenumbers]{cvpr} 

\usepackage{graphicx}
\usepackage{amsmath}
\usepackage{amssymb}
\usepackage{booktabs}

\usepackage{bm}
\usepackage{tabularx}
\usepackage[linesnumbered, ruled]{algorithm2e}

\newcolumntype{C}{>{\centering\arraybackslash}X}
\newcolumntype{S}{>{\hsize=.3\hsize}>{\centering\arraybackslash}X}
\newcolumntype{T}{>{\hsize=.7\hsize}>{\centering\arraybackslash}X}

\usepackage[pagebackref,breaklinks,colorlinks]{hyperref}

\usepackage[capitalize]{cleveref}
\crefname{section}{Sec.}{Secs.}
\Crefname{section}{Section}{Sections}
\Crefname{table}{Table}{Tables}
\crefname{table}{Tab.}{Tabs.}

\def\cvprPaperID{****} 
\def\confName{CVPR}
\def\confYear{****}

\begin{document}

\title{Supervised Contrastive Learning on Blended Images for Long-tailed Recognition}

\author{Minki Jeong \quad Changick Kim\\
Korea Advanced Institute of Science and Technology\\
{\tt\small \{rhm033, changick\}@kaist.ac.kr}
}
\maketitle

\begin{abstract}
Real-world data often have a long-tailed distribution, where the number of samples per class is not equal over training classes. The imbalanced data form a biased feature space, which deteriorates the performance of the recognition model. In this paper, we propose a novel long-tailed recognition method to balance the latent feature space. First, we introduce a MixUp-based data augmentation technique to reduce the bias of the long-tailed data. Furthermore, we propose a new supervised contrastive learning method, named Supervised contrastive learning on Mixed Classes (SMC), for blended images. SMC creates a set of positives based on the class labels of the original images. The combination ratio of positives weights the positives in the training loss. SMC with the class-mixture-based loss explores more diverse data space, enhancing the generalization capability of the model. Extensive experiments on various benchmarks show the effectiveness of our one-stage training method.
\end{abstract}

\section{Introduction}
\label{SMCsec:intro}

Real-world data could be long-tail distributed, where the number of samples of each class changes drastically.
This imbalance is due to various reasons, such as data collecting costs or lack of data.
Training a neural network with imbalanced data is likely to produce an overfitted network biased toward major classes (\ie, head classes) with poor recognition performances in minor classes (\ie, tail classes).

\begin{figure}
    \centering
    \includegraphics[width=0.95\linewidth]{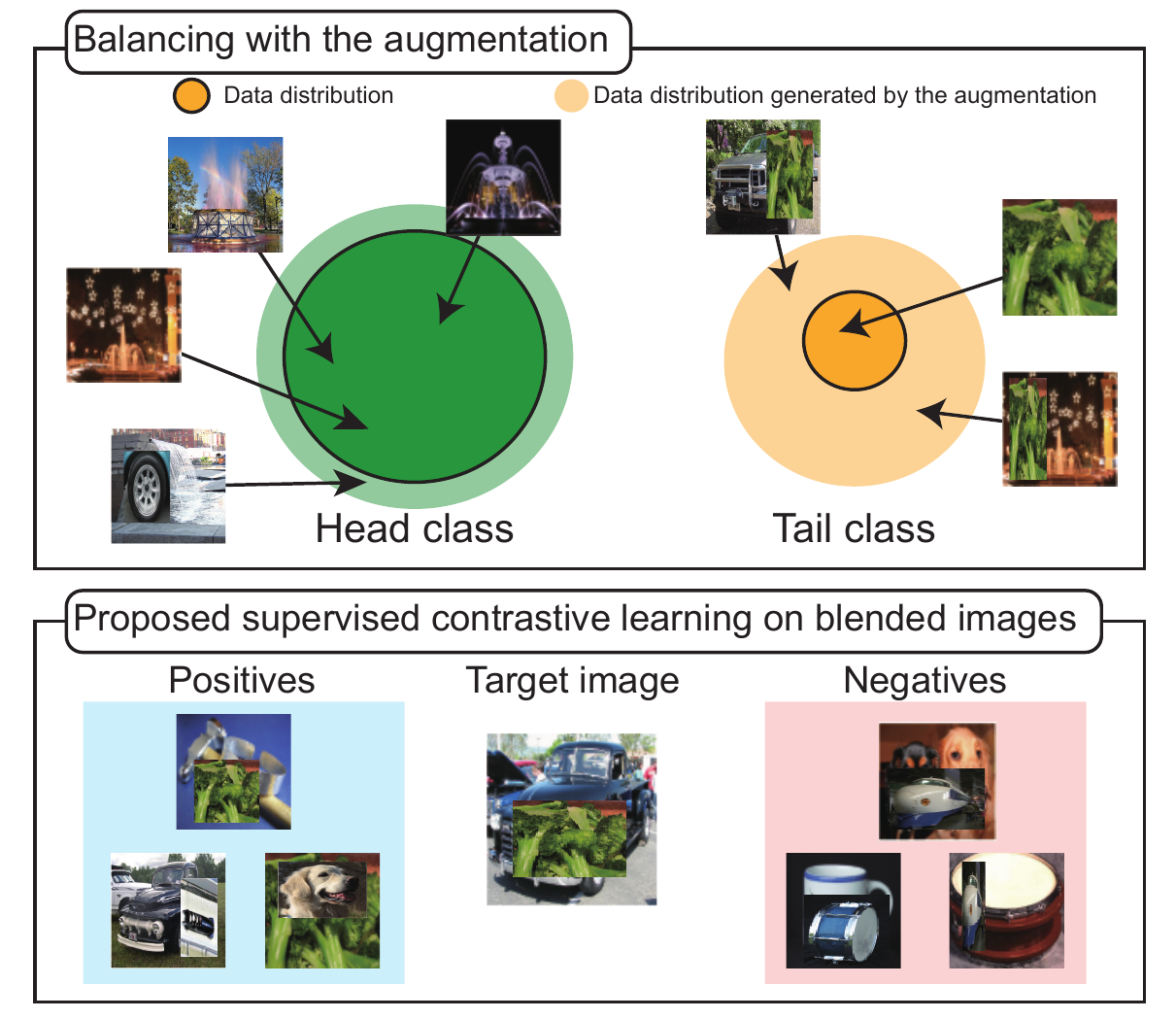}
    \caption{The concept image of SMC. Data augmentation reduces the imbalance between head and tail classes (fountain and broccoli in this example) without strong regularization. Moreover, our supervised contrastive learning method on blended images improves recognition performances.
    }
    \label{SMCfig:frontal}
\end{figure}

Re-sampling and re-weighting methods have been major approaches to the data imbalance problem.
Re-sampling methods~\cite{chawla2002smote, buda2018systematic, more2016survey, zhang2021bag} are artificially balance the training data by assigning different data sampling weights to the training data. When creating a mini-batch for training, those methods oversample tail class instances or undersample head class instances.
However, re-sampling approaches could worsen the overfitting issue due to the lack of semantic diversity of the sampled data.
Re-weighting methods~\cite{cao2019learning, cui2019class, ren2020balanced, lin2017focal, li2022equalized} modify the training loss for the imbalanced condition. They assign different weights on training instances to adjust the impact of training samples on the model.
Although re-weighting enhances the recognition performance on the tail class samples, relatively reduced weights on head class samples discourage the head class sample recognition~\cite{kang2019decoupling}.

Several recent studies utilize the generalization capability of supervised contrastive learning (SCL) to long-tailed recognition~\cite{KCL, TSC}.
It is shown that SCL increases the generalization capability in various applications. Whereas the SCL approach could enhance the feature extractor in the long-tailed condition, it often generates a head-biased feature space due to the nature of long-tailed distribution~\cite{KCL}.
Several long-tailed recognition studies~\cite{KCL, TSC, BCL} introduce regularization approaches to balance the feature space.
However, the strong regularization could interfere with the class representation capability of the feature extractor, which deteriorates its recognition performance.

\begin{figure}[t]
    \centering
     \includegraphics[width=0.95\linewidth]{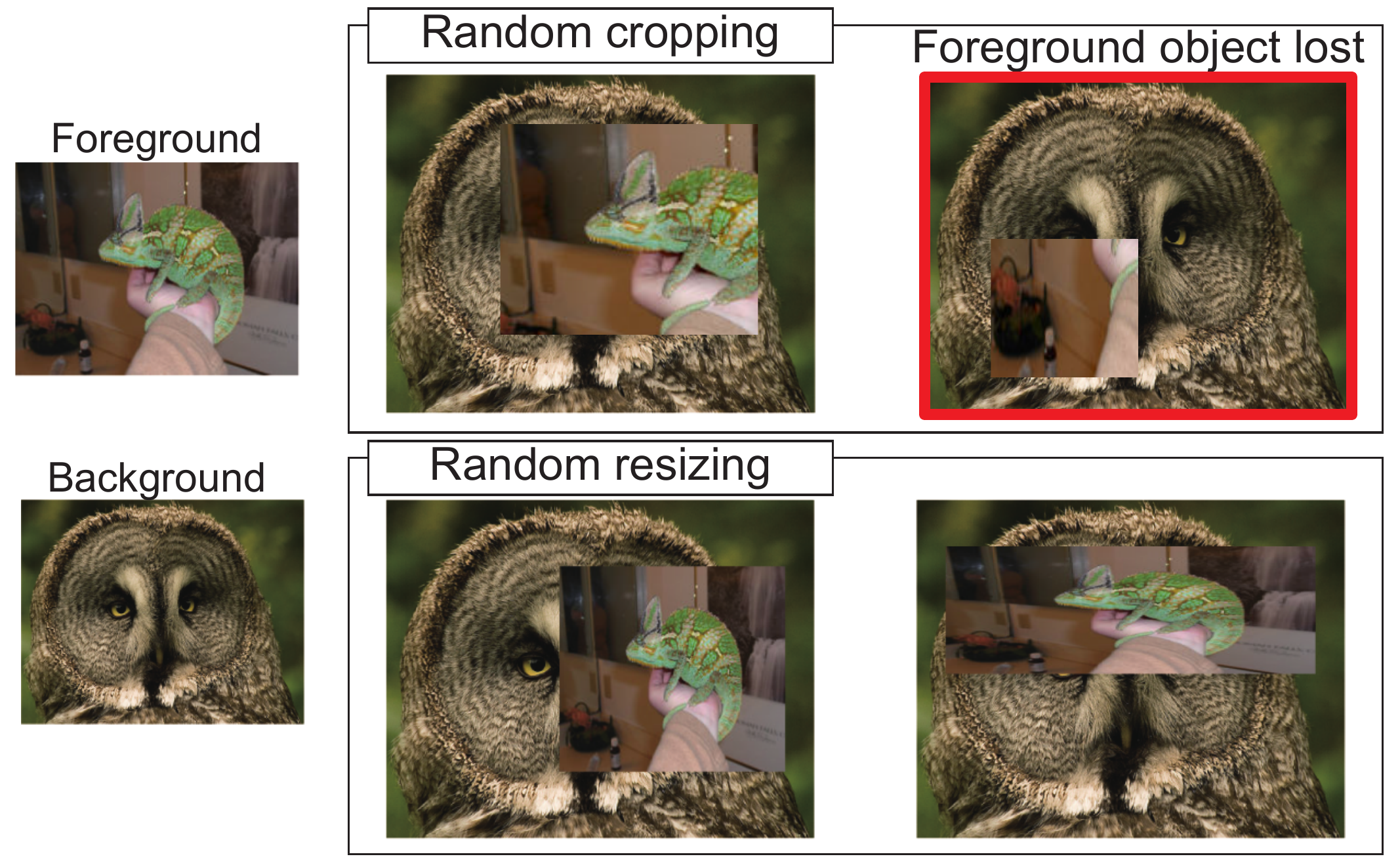}
    \caption{Comparison of the two data augmentation methods. Due to the cropping operation, the semantic information of foreground images could be discarded (chameleon in this example). We replace the cropping with resizing to preserve the semantic information.}
    \label{SMCfig:mixup_comparison}
\end{figure}

In this paper, we propose a novel long-tailed recognition method that forms a balanced feature space with SCL. The main concept of our method is to reduce the class imbalance by mixing the training samples without strong regularization.
The data mixing increases the semantic diversity of tail class information during training. Moreover, the randomly sampled coefficient for the data augmentation expands the available semantic information space for training, as illustrated in \cref{SMCfig:frontal}.
The broader semantic space gives richer information to the model than using a set of pre-defined class data, which encourages the classification power of the model.
Among the various data mixing methods, CutMix~\cite{CutMix} shows good performance in long-tailed recognition~\cite{CMO}. However, the random cropping operation in CutMix may lead to loss of foreground class information, as illustrated in \cref{SMCfig:mixup_comparison}. Therefore, we use ResizeMix~\cite{ResizeMix} to preserve the semantic information of the foreground class. Since ResizeMix omits random cropping for the foreground patch, the foreground knowledge is preserved.
Furthermore, we introduce a new SCL method that considers the semantic information of the original classes, named Supervised contrastive learning on Mixed Classes (SMC), for long-tailed recognition.
The mixed data contain multiple class features. Therefore, the positive and negative pairs should be specified for the mixed data to train with supervised contrastive approach.
We define three categories of positive pairs with mixed data.
Then, we assign weights to the positive pairs based on the combination ratio of mixed samples. The weighted loss function is used to train the network.
Moreover, we propose a new data augmentation method for supervised contrastive learning of the class-mixed samples. 
We found that random cropping on the blended data to create positive pairs could lose the semantic information of the training samples.
Based on our observation, we propose a new data augmentation method that preserves the original class information of the mixed sample. Instead of applying augmentations on mixed images, we augment the images before mixing. Then, we create blended images with the augmented images. Our approach retains the information in mixtures which leads to better performance.

We evaluate SMC on the three long-tailed recognition benchmarks: CIFAR-100-LT~\cite{CIFAR-LT}, ImageNet-LT~\cite{ImageNet-LT}, and iNaturalist 2018~\cite{iNaturalist}. Our method achieves state-of-the-art performances on the benchmarks. In addition, we provide a deep analysis of SMC that justifies the concept of our method.

To summarize, our contribution is three-fold:
\begin{itemize}
    \item We suggest a new regularization method to reduce the class imbalance for long-tailed recognition. Using blended data provides more diversity to the network, increasing the generalization capability of the model.
    \item We propose a novel supervised contrastive learning method, named Supervised contrastive learning on Mixed Classes (SMC), for long-tailed recognition. The data mixtures encourage the generalization capability of the model. In addition, we suggest positive and negative pair definitions and a new data augmentation method for SMC. Our proposals maintain the original class information of a mixed sample to balance the feature space.
    \item We provide an in-depth experimental analysis of SMC. Our method achieves state-of-the-art performance on various long-tailed recognition benchmarks. Moreover, empirical evaluations on our proposals justify the contribution of our method.
\end{itemize}

\section{Related work}
\label{SMCsec:relwork}
\subsection{Long-tailed recognition}
Re-sampling methods create a subset of training data that consists of class-balanced samples.
Oversampling~\cite{chawla2002smote, buda2018systematic} or undersampling~\cite{more2016survey, zhang2021bag} methods have been proposed to achieve the class balance of the subset.
Oversampling approaches match the number of samples per class by repeatably collecting tail class instances.
On the contrary, undersampling methods omit head class samples to equalize the label distribution.
Training with the balanced subset reduces defects caused by the class imbalance.
However, the lack of diversity of the tail class data still needs to be solved, which endangers the network to overfitting. Re-weighting methods calculate weights of training samples for the training loss function.
The weights could be derived from the label distribution~\cite{cao2019learning, cui2019class, ren2020balanced}, or instance-level scoring functions~\cite{lin2017focal, li2022equalized}.
The weighted loss decreases the dominance of head information, which leads to the trade-off between the head and tail class recognition performances.
Using multiple experts~\cite{RIDE, li2022trustworthy} also shows good performance. The experts compensate each other to boost the recognition performance.
More recently, Mixup-based data augmentation has attracted attention in the field of long-tailed recognition. Remix~\cite{chou2020remix} introduces a class distribution-aware labeling for the image mixtures. Park~\etal~\cite{CMO} suggest a MixUp-based data augmentation method, named CMO, to increase the information diversity of head-biased data. The variance-enhanced training data provokes a more generalized feature space, leading to better recognition performances.
Our method is in line with Remix and CMO in terms of using data augmentation to adjust the biased distribution of training data. The difference is that SMC utilizes the generalization capability of SCL on top of the mixtures.

\subsection{Supervised contrastive learning on long-tailed recognition}
Supervised contrastive learning~\cite{khosla2020supervised} utilizes class labels to define positive and negative pairs of contrastive learning.
The positive pairs get closer, while the negative pairs get farther away.
SCL shows good performance in various applications~\cite{khosla2020supervised, liu2021learning, Zhao_2022_CVPR}.
However, in long-tailed recognition, using SCL directly causes a head-biased feature space and a biased classifier that could hinder the network~\cite{KCL}. To reduce the bias, KCL~\cite{KCL} screens the positive pairs during training. Li~\etal~\cite{TSC} introduce TSC, a two-stage learning method with regularization on the uniformly pre-defined class center points.
BCL~\cite{BCL} utilizes the classifier weights as the additional class information for SCL.
Our concept is inspired by the feature regularization approaches mentioned above.
The difference is that our method balances the feature space using augmented data without strong regularization.

\subsection{MixUp-based data augmentation}
MixUp~\cite{MixUp} is proposed for more robust training of neural networks. It shows that using mixed instances is in line with the risk minimization of neural networks.
Yun~\etal~\cite{CutMix} claim that the linear combination of images in MixUp creates unnatural and ambiguous images that could deteriorate the performance of networks. They propose CutMix, which replaces the linear combination of images in MixUp with random cropping and pasting to create locally meaningful images.
Qin~\etal~\cite{ResizeMix} point out the possible information loss of the random cropping in CutMix.
They introduce ResizeMix, which uses image resizing rather than the random cropping.
The high generalization capability of MixUp-based augmentations boosts various computer vision applications~\cite{mangla2020charting, CMO}.
Recently, SDMP~\cite{Ren_2022_CVPR} has utilized MixUp in self-supervised learning. The main concept of SDMP is using ResizeMix to create positives for self-supervised contrastive learning, similar to our idea.
The difference between SDMP and SMC is that our method aims at supervised contrastive learning with class labels on long-tailed recognition, while SDMP focuses on self-supervised learning.

\section{Propose method}
\label{SMCsec:method}

\subsection{Class imbalance and data mixing}
The imbalance of the latent feature space is the main challenge to use SCL in long-tailed recognition. Applying SCL without considering class imbalance results in a head-biased feature space, which discourages tail class recognition. Previous studies add regularization to the network to balance the feature space.
However, we suspect that the strong regularization could lead to suboptimal results since tail classes still have small variances.
Therefore, we use the MixUp-based~\cite{MixUp} data augmentation method to expand the variety of training data. As illustrated in \cref{SMCfig:frontal}, the interpolated data form a broader data space than the original data, which increases the generalization capability of the network. Furthermore, using the blended data could relax the information imbalance of the long-tailed data.

Among the variations of MixUp, CutMix~\cite{CutMix} shows a good performance in the previous long-tailed recognition study~\cite{CMO}.
However, the random cropping nature of CutMix may lead to loss of the critical information of foreground classes as illustrated in \cref{SMCfig:mixup_comparison}.
Although the information-removed samples could prevent overfitting, the information loss is undesirable since CutMix assumes the presence of multiple class features in the blended image.
Thus, we replace the random cropping with a resizing operation to secure the foreground class information, which leads to better generalization capability of the trained model.
For foreground and background images $x_{f}$ and $x_{b}$ and their labels $y_{f}$ and $y_{b}$, the mixed image $\tilde{x}$ and its label $\tilde{y}$ are defined as follows:
\begin{equation}
\begin{aligned}
&\tilde{x} = \bm{M} \odot R(x_{f}) + (\bm{1} - \bm{M}) \odot x_{b},\\
&\tilde{y} = \lambda y_{f} + (1 - \lambda) y_{b}, 
\end{aligned}
\label{SMCeq:resizemix}
\end{equation}
where $\bm{M} \in \{0, 1\}^{W \times H}$ is a binary mask for the images, $R(x_{f})$ is a resized and padded foreground, $\odot$ implies the Hadamard product and $\lambda \sim Beta( \alpha, \alpha )$ indicates the combination ratio that sampled from the beta distribution.
We normalized the combination ratio in the range of $[0.2, 0.8]$ to increase the diversity of training data by preserving both class features in the mixed output.

\begin{figure*}[t]
    \centering
    \includegraphics[width=0.95\linewidth]{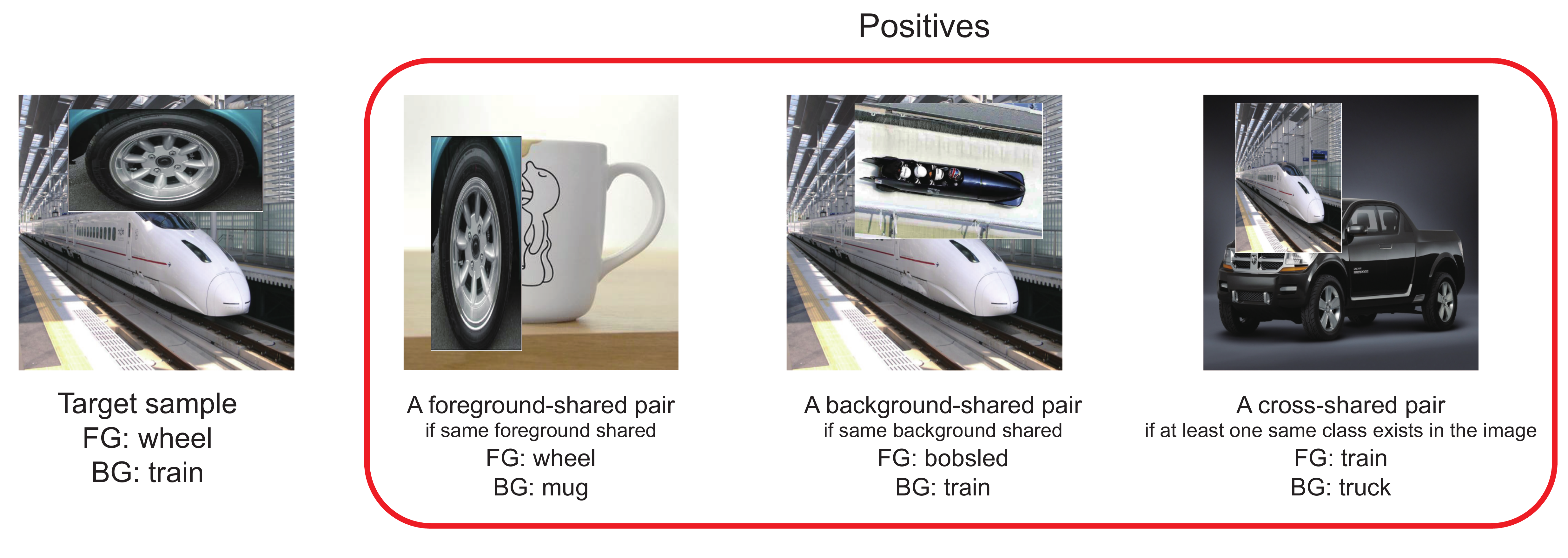}
    \caption{Three positive types of SMC.
    If a mixed image contains an image belonging to the same class, we define it as a positive pair.
    The foreground of the foreground-shared pair shares the same class, whereas the background of the background-shared pair shares the same class. A cross-shared pair is an image that is neither a foreground-shared pair nor a background-shared pair but shares at least one same class. The positives help the network reduce the imbalance of its feature space.}
    \label{SMCfig:pos_neg_definition}
\end{figure*}

The next issue is the way to sample the foregrounds and backgrounds. On long-tailed data, the uniform sampling could lead to oversampling of head samples, which is undesirable in terms of data diversity.
Inspired by the previous study~\cite{CMO}, we randomly sample background images with a uniform probability, while foreground samples have a weighted sampler that samples the tail class instances more frequently. The weighted sampling for foregrounds balances between head and tail class information in the mixture distribution. The sampling probability for the $k$-th class is defined as follows:
\begin{equation}
    q(k) = \frac{n_{k}^{-\gamma}}{\sum_{l=1}^{C}n_{l}^{-\gamma}},
    \label{SMCeq:sampling_prob}
\end{equation}
where $n_{k}$ is the number of samples in the $k$-th class, $C$ implies the number of training classes and $\gamma$ is a hyperparameter that determines the sampling strategy. We set $\gamma$ to one based on the previous study~\cite{CMO}.

\subsection{Supervised contrastive learning on mixed data}
SCL requires to define positives and negatives for training samples. Here, positives share the same class label, and negatives have samples with different labels from the target sample.
However, this definition is inappropriate for the mixed data because a blended sample contains multiple class information.
Therefore, we suggest three positive pair types as illustrated in \cref{SMCfig:pos_neg_definition}: foreground-shared pairs, background-shared pairs, and cross-shared pairs.
A foreground-shared and background-shared pairs share the foreground and background classes in a mixed image, respectively.
A cross-shared pair is an image that shares at least one same class, but is neither a foreground-shared pair nor a background-shared pair.
The negative set consists of the remaining samples except for the positives.

\begin{figure}[t]
    \centering
    \includegraphics[width=0.95\linewidth]{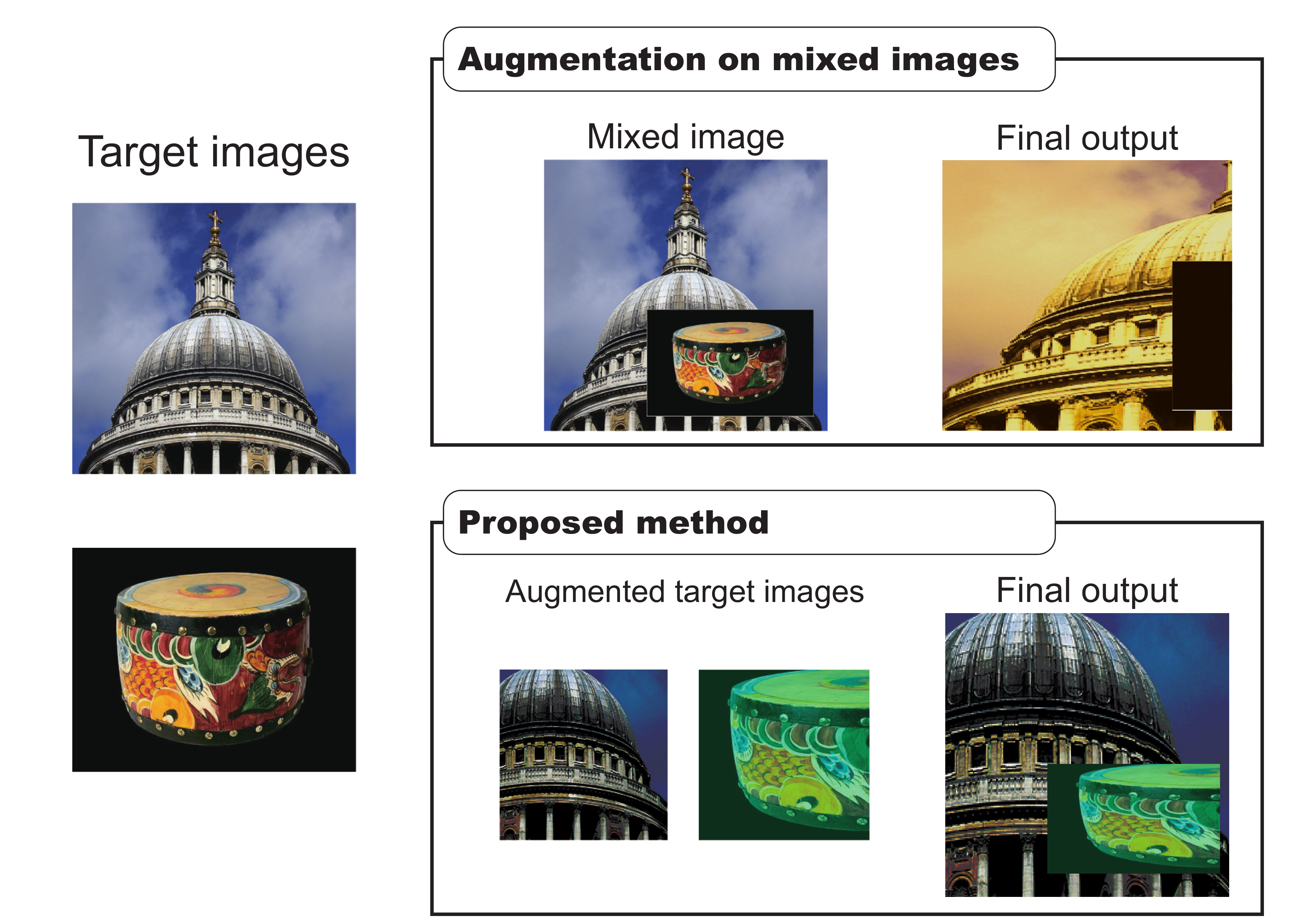}
    \caption{Data augmentation on the mixtures. Since the random cropping on mixed images could remove the semantic information of the original images (drum in this example), we augment the component images before mixing to preserve both class features.}
    \label{SMCfig:mixture_augmentation}
\end{figure}

Previous contrastive learning methods augment the target image to produce positives~\cite{chen2020simple, khosla2020supervised}. Among the various augmentation operations, random cropping is a crucial factor in enhancing the generalization capability~\cite{chen2020simple}. However, the random cropping on the mixture is in danger of the information loss of the original images as illustrated in \cref{SMCfig:mixture_augmentation}.
Therefore, we suggest adjusting the original images before blending to secure the diversity for contrastive learning:
\begin{equation}
\begin{aligned}
&\tilde{x} = \bm{M} \odot R(\tilde{x}_{f}) + (\bm{1} - \bm{M}) \odot \tilde{x}_{b},\\
&\tilde{y} = \lambda y_{f} + (1 - \lambda) y_{b},
\end{aligned}
\label{SMCeq:data_aug}
\end{equation}
where $\tilde{x}_{f}$ and $\tilde{x}_{b}$ are the augmented foreground and background images, respectively.
This approach preserves the information of foreground classes while benefits from the rich generalization capability of image mixing.

\begin{figure*}[t]
    \centering
    \includegraphics[width=0.9\textwidth]{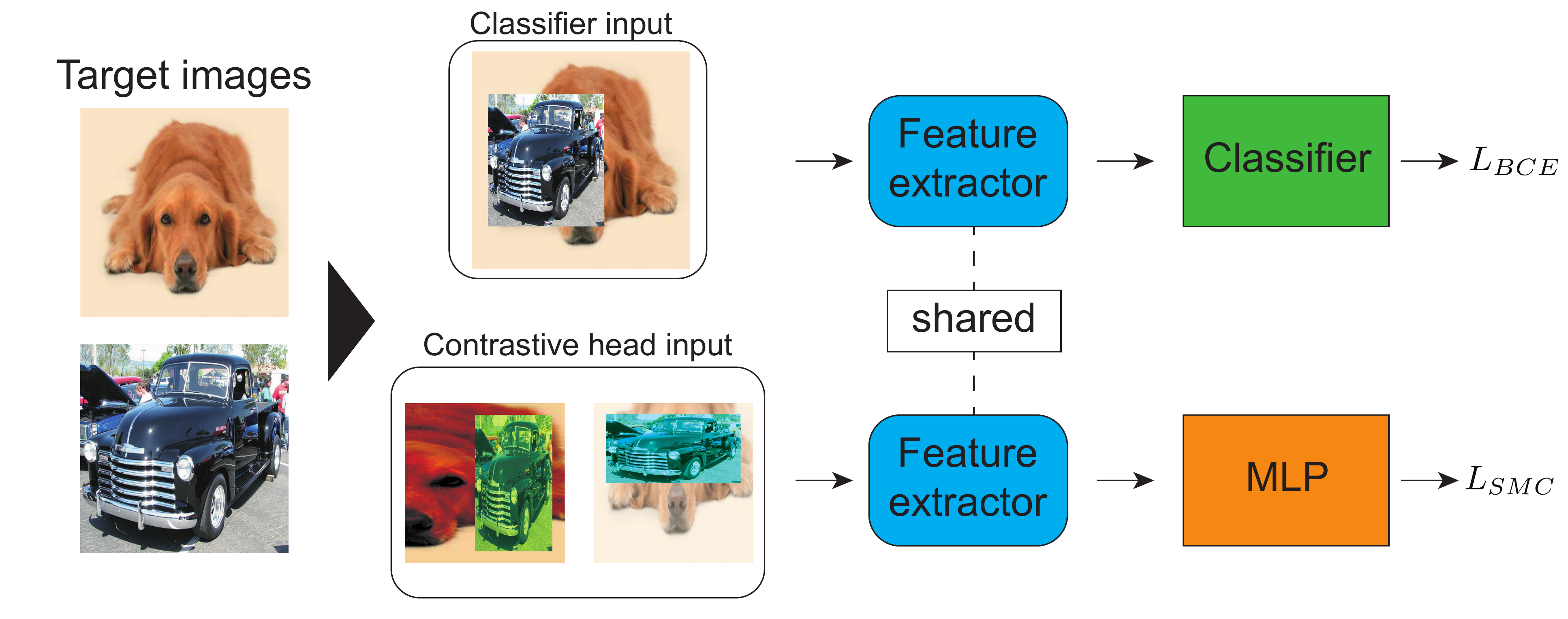}
    \caption{Illustration of SMC. The MixUp-based augmentation creates more diverse data. The blended data are used for training the classifier and the MLP head with the corresponding positives and negatives.}
    \label{SMCfig:overall}
\end{figure*}

Finally, the SMC loss $L_{SMC}$ on a set of features of augmented images $\tilde{x}_{i} \in \tilde{\mathcal{X}}$ is defined as follows:
\begin{equation}
\begin{aligned}
    &L_{f} = -\frac{1}{|\tilde{\mathcal{X}}|}\sum_{\tilde{x}_{i} \in \tilde{\mathcal{X}}}{\frac{1}{|\tilde{\mathcal{X}}_{i}^{f}|}\sum_{\tilde{x}_{f} \in \tilde{\mathcal{X}}_{i}^{f}}{\log{\frac{\exp{(\tilde{x}_{i} \tilde{x}_{f}/\tau)}}{\sum_{\tilde{x}_{j} \in \tilde{\mathcal{X}}_{i}}{\exp{(\tilde{x}_{i}\tilde{x}_{j}/\tau)}}}}}},\\
    &L_{b} = -\frac{1}{|\tilde{\mathcal{X}}|}\sum_{\tilde{x}_{i} \in \tilde{\mathcal{X}}}{\frac{1}{|\tilde{\mathcal{X}}_{i}^{b}|}\sum_{\tilde{x}_{b} \in \tilde{\mathcal{X}}_{i}^{b}}{\log{\frac{\exp{(\tilde{x}_{i} \tilde{x}_{b}/\tau)}}{\sum_{\tilde{x}_{j} \in \tilde{\mathcal{X}}_{i}}{\exp{(\tilde{x}_{i}\tilde{x}_{j}/\tau)}}}}}},\\
    &L_{c} = -\frac{1}{|\tilde{\mathcal{X}}|}\sum_{\tilde{x}_{i} \in \tilde{\mathcal{X}}}{\frac{1}{|\tilde{\mathcal{X}}_{i}^{c}|}\sum_{\tilde{x}_{c} \in \tilde{\mathcal{X}}_{i}^{c}}{\log{\frac{\exp{(\tilde{x}_{i} \tilde{x}_{c}/\tau)}}{\sum_{\tilde{x}_{j} \in \tilde{\mathcal{X}}_{i}}{\exp{(\tilde{x}_{i}\tilde{x}_{j}/\tau)}}}}}},\\
    &L_{SMC} = w_{f} L_{f} + w_{b} L_{b} + w_{c} L_{c},
\end{aligned}
\label{SMCeq:smc_loss}
\end{equation}
where $\tilde{\mathcal{X}}_{i} = \tilde{\mathcal{X}} \setminus {\tilde{x}_{i}}$ is a set of the mini-batch features without $\tilde{x}_{i}$, $|\mathcal{X}|$ indicates the mini-batch size.
$\tilde{\mathcal{X}}_{i}^{f}$, $\tilde{\mathcal{X}}_{i}^{b}$, and $\tilde{\mathcal{X}}_{i}^{c}$ are a set of foreground-shared pairs, background-shared pairs, and cross-shared pairs, respectively. $|\tilde{\mathcal{X}}_{i}^{f}|$, $|\tilde{\mathcal{X}}_{i}^{b}|$, and $|\tilde{\mathcal{X}}_{i}^{c}|$ indicate set sizes of the corresponding positives, and $\tau$ is a temperature parameter set to $0.1$.
The loss weights $w_{f} = \lambda_{w} / 1.5$, $w_{b} = (1 - \lambda_{w}) / 1.5$, $w_{c} = 0.5 / 1.5$ follow $\lambda_{w}$, the combination ratio of images in the pair.
The weights are normalized to make the total summation of the weights to one.
They represent the semantic relationship between images in a pair.
Positive pairs with stronger relationships have a greater impact on the network than pairs with smaller relationships.

\subsection{Classification loss on mixed data}
The cross-entropy function is a typical choice to train classifiers. In the long-tailed environment, however, it often results in a head-biased classifier due to imbalanced training data~\cite{ren2020balanced}.
The logit compensation methods~\cite{CIFAR-LT, menon2020long} aim to reduce the bias by adjusting the classifier output with the knowledge on the class distribution.
We use the compensation method to balance the classifier with the mixed data.
Since the mixing augmentation increases the data variance, the mixtures aid the classifier to explore more broader data space.
Furthermore, training the classifier and feature extractor simultaneously achieves the one-stage training of the model.
With the blended data, the classifier is trained with the following loss function:
\begin{equation}
L_{BCE} = CE(f(\tilde{x}) + \bm{m}, \tilde{y}),
\label{SMCeq:bce_loss}
\end{equation}
where $CE(\cdot, \cdot)$ is the cross-entropy function, and $\bm{m} = \log{(\mathbb{P}_{Y})}$ is a prior vector of the class distribution $\mathbb{P}_{Y}$ with the logarithm function for the compensation.
The prior vector gives higher values for head classes, which emphasizes tail class instances during training.

By combining the losses defined above, the total training loss is defined as follows:
\begin{equation}
L_{train} = L_{BCE} + \eta L_{SMC},
\label{SMCeq:total_loss}
\end{equation}
where $\eta$ is a weight hyperparameter.
\Cref{SMCfig:overall} illustrates the summary of SMC.

\section{Experimental analysis}
\label{sec:experiments}

\subsection{Datasets and comparison metrics}
We used CIFAR-100-LT~\cite{CIFAR-LT}, ImageNet-LT~\cite{ImageNet-LT}, and iNaturalist 2018~\cite{iNaturalist} to evaluate SMC. The datasets have different imbalance ratios. The imbalance ratio $\rho$ is defined as the ratio of the number of instances in the largest head class to that of the smallest tail class.
CIFAR-100-LT is a subset of the CIFAR-100 dataset~\cite{CIFAR}. It artificially excludes training data to make 100 long-tail distributed classes. We used three imbalance ratios for comparisons: 100, 50, and 10.
ImageNet-LT is a subset of the ImageNet 2012 dataset~\cite{ImageNet}. Similar to CIFAR-100-LT, it selects training data to create a set of 1000 imbalanced classes. There are $115.8\text{K}$ images and its imbalance ratio is 256.
On the contrary to the previous datasets, the iNaturalist 2018 dataset is imbalanced from the data collecting stage.
There are $437,513$ training images of $8,142$ classes, where the imbalance ratio is 500.

\subsection{Implementation details}
We followed the training strategy of previous long-tailed recognition studies~\cite{KCL, TSC, CMO} for a fair comparison. we used the ResNet-32~\cite{resnet} backbone on CIFAR-100-LT and the ResNet-50~\cite{resnet} backbone on ImageNet-LT and iNaturalist 2018.
On CIFAR-100-LT, we followed the training strategy in \cite{cao2019learning}.
For ImageNet-LT and iNaturalist 2018, the initial learning rate is set to $0.1$. In the ImageNet-LT case, the number of training epochs is set to 100 and the learning rate is decreased by $0.1$ at epoch 60 and 80. The number of samples in a mini-batch is 128.
In the iNaturalist 2018 case, we trained the network 200 epochs and we reduced the learning rate at epoch 75 and 160 by $0.1$.
The training mini-batch size is set to 256 for the iNaturalist 2018 dataset.
For all datasets, we set the loss weight $\eta$ to $0.1$ and the output dimension of the 2-layer MLP for contrastive learning to 128. The $\alpha$ value for the Beta distribution is set to $1.0$, and the hidden layer size of the head follows the input feature size.
Please see the supplementary material for further details.

\subsection{Comparison with the other methods}
Since SMC utilizes the power of SCL and MixUp-based augmentation, we select SCL-based methods~\cite{KCL, TSC, BCL} and MixUp-based methods~\cite{chou2020remix, TSC} with previous long-tailed recognition methods~\cite{kang2019decoupling, cui2019class, lin2017focal, cao2019learning} as the comparison baselines.
The comparison result on CIFAR-100-LT is in \Cref{SMCtable:cifar_comparison}.
SMC shows better performances than baselines in all comparisons.
This result implies that SMC is effective for various long-tailed conditions.
The mixed samples increase the semantic information space of training data, which improves the discriminative power of the network. Furthermore, the blended data gives a soft balancing regularization to the network. The soft regularization reduces the bias in the feature space that leads to better recognition capability.

The performance comparisons on the ImageNet-LT and iNaturalist 2018 datasets are in \Cref{SMCtable:imagenet_comparison} and \Cref{SMCtable:inat_comparison}, respectively.
The overall recognition performance of SMC is still better than the baselines.
This comparison result demonstrates that SMC is still effective when the data distribution is complex.
Compared with TSC, an SCL-based two-stage learning method, SMC has better performance on medium and few shot classes, while the head class performance is weaker than TSC.
It shows that the broader training data space and soft regularization are helpful for low-shot classes, but harms the classification on major classes. Nevertheless, the performance gain on low-shot classes significantly improves the overall recognition capability. We leave the performance balancing between many-shot and low-shot classes as an open problem.

\begin{table}[t]
    \centering
    \begin{tabularx}{0.95\linewidth}{>{\hsize=.35\hsize}X | >{\hsize=.15\hsize}>{\centering\arraybackslash}X | >{\hsize=.15\hsize}>{\centering\arraybackslash}X | >{\hsize=.15\hsize}>{\centering\arraybackslash}X}
         \hline
         Method &  $\rho = 100$ & $\rho = 50$ & $\rho = 10$ \\
         \hline
         CE                 & $38.3$ & $43.9$ & $55.7$ \\
         BS~\cite{cui2019class} & $38.6$ & $44.6$ & $57.1$ \\
         Focal~\cite{lin2017focal}        & $38.4$ & $44.3$ & $55.8$ \\
         BS-Focal~\cite{cui2019class}     & $39.6$ & $45.2$ & $58.0$ \\
         CE-DRW~\cite{cao2019learning}    & $40.5$ & $44.7$ & $56.2$ \\
         CE-DRS~\cite{cao2019learning}    & $40.4$ & $44.5$ & $56.1$ \\
         LDAM~\cite{cao2019learning}  & $39.6$ & $45.0$ & $56.9$ \\
         LDAM-DRW~\cite{cao2019learning}  & $42.0$ & $46.2$ & $58.7$ \\
         Remix~\cite{chou2020remix} & $45.8$ & $49.5$ & $59.2$ \\
         CMO~\cite{CMO}     & $46.6$ & $51.4$ & $62.3$ \\
         KCL~\cite{KCL}     & $42.8$ & $46.3$ & $57.6$ \\
         TSC~\cite{TSC}     & $43.8$ & $47.4$ & $59.0$ \\
         Ours (SMC)         & $\bm{48.9}$ & $\bm{52.3}$ & $\bm{62.5}$ \\
         \hline
    \end{tabularx}
    \caption{Classification accuracy (\%) on the CIFAR-100-LT dataset. Please see the supplementary material for more details.}
    \label{SMCtable:cifar_comparison}
\end{table}

\begin{table}[t]
    \centering
    \begin{tabularx}{0.95\linewidth}{>{\hsize=.25\hsize}X | >{\hsize=.1\hsize}>{\centering}X  >{\hsize=.15\hsize}>{\centering}X  >{\hsize=.1\hsize}>{\centering}X | >{\hsize=.12\hsize}>{\centering\arraybackslash}X}
         \hline
         Method &  Many & Medium & Few & All \\
         \hline
         $\tau$-norm~\cite{kang2019decoupling}    & $56.6$ & $44.2$ & $27.4$ & $46.7$ \\
         cRT~\cite{kang2019decoupling}            & $58.8$ & $44.0$ & $26.1$ & $47.3$ \\
         LWS~\cite{kang2019decoupling}            & $57.1$ & $45.2$ & $29.3$ & $47.7$ \\
         FCL~\cite{KCL} & $61.4$ & $47.0$ & $28.2$ & $49.8$ \\
         Remix~\cite{chou2020remix} & $60.4$ & $46.9$ & $30.7$ & $48.6$\\
         CMO~\cite{CMO} & $62.0$ & $49.1$ & $36.7$ & $52.3$ \\
         KCL~\cite{KCL} & $61.8$ & $49.4$ & $30.9$ & $51.5$ \\
         TSC~\cite{TSC} & $\bm{63.5}$ & $49.7$ & $30.4$ & $52.4$ \\
         Ours (SMC)     & $61.8$ & $\bm{49.8}$ & $\bm{37.9}$ & $\bm{52.7}$ \\
         \hline
         \textdagger BCL\cite{BCL}     & $-$ & $-$ & $-$ & $56.0$ \\
         \textdagger Ours (SMC)     & $66.9$ & $54.2$ & $36.1$ & $\bm{56.6}$ \\
         \hline
    \end{tabularx}
    \caption{Classification accuracy (\%) on the ImageNet-LT dataset. \textdagger~denotes results with the different training setup, presented in \cite{BCL}. Please see the supplementary material for further details.}
    \label{SMCtable:imagenet_comparison}
\end{table}

\begin{table}[t]
    \centering
    \begin{tabularx}{0.95\linewidth}{>{\hsize=.25\hsize}X | >{\hsize=.1\hsize}>{\centering}X  >{\hsize=.15\hsize}>{\centering}X  >{\hsize=.1\hsize}>{\centering}X | >{\hsize=.12\hsize}>{\centering\arraybackslash}X}         \hline
         Method &  Many & Medium & Few & All \\
         \hline
         CE             & $72.2$ & $63.0$ & $57.2$ & $61.7$ \\
         cRT~\cite{kang2019decoupling}            & $69.0$ & $66.0$ & $63.2$ & $65.2$ \\
         $\tau$-norm~\cite{kang2019decoupling}    & $65.6$ & $65.3$ & $65.9$ & $65.6$ \\
         LWS~\cite{kang2019decoupling}            & $65.0$ & $66.3$ & $65.5$ & $65.9$ \\
         Remix~\cite{chou2020remix} & $-$    & $-$    & $-$ & $70.5$ \\
         \textdagger CMO~\cite{CMO} & $67.7$ & $69.3$ & $70.7$ & $69.7$ \\
         KCL~\cite{KCL} & $-$    & $-$    & $-$    & $68.6$ \\
         TSC~\cite{TSC} & $\bm{72.6}$ & $\bm{70.6}$ & $67.8$ & $69.7$ \\
         Ours (SMC)     & $69.6$ & $70.1$ & $\bm{71.3}$ & $\bm{70.6}$\\
         \hline
    \end{tabularx}
    \caption{Classification accuracy (\%) on the iNaturalist 2018 dataset. \textdagger~denotes the result with our training setup. Please see the supplementary material for more details.}
    \label{SMCtable:inat_comparison}
\end{table}

\subsection{Feature space analysis with contrastive loss}
\label{SMCsec:experiments_feature_space}
In this subsection, we evaluate the contribution of SMC by analyzing the feature space created with SMC.
We use two metrics to assess the trained feature space: Inter-class and Semantic similarity scores.

Inter-class score $IS_{k}$ for the $k$-th class is defined as follows:
\begin{equation}
IS_{k} = \exp \left( -\frac{1}{C}\sum_{j=1}^{C}{\left( \bm{c}_{k} - \bm{c}_{j} \right) / \tau'} \right),
\label{SMCeq:inter_class_score}
\end{equation}
where $C$ is the number of classes, $\bm{c}_{k}$ is a class feature vector calculated from an average of the feature vectors of the $k$-th class, and $\tau'$ is a temperature value to scale score values. We set $\tau'$ to 10 for comparisons.
The lower inter-class score means that the class centers are far apart from each other, which implies that the network can distinguish different classes more easily.

The well-trained feature space contains the semantic information of the training classes. If two classes are semantically similar with each other, they will be placed close to each other within the space.
Semantic similarity score measures the similarity between class semantic information and the latent space. It measures the difference between cosine similarities of class semantic vectors and that of the class feature vectors.
Semantic similarity score $SS$ is defined as follows:
\begin{equation}
\begin{aligned}
&SS = \frac{1}{C^{2}} \sum_{i=1}^{C}{\sum_{j=1}^{C}{||\mathbf{S}_{i,j}^{s} - \mathbf{S}_{i,j}^{c}||_{1}}},\\
&\mathbf{S}^{s}_{i,j} = \frac{\bm{s}_{i}\bm{s}_{j}}{||\bm{s_}{i}||_{2} \cdot ||\bm{s}_{j}||_{2}} \quad \mathbf{S}^{c}_{i,j} = \frac{\bm{c}_{i}\bm{c}_{j}}{||\bm{c}_{i}||_{2} \cdot ||\bm{c}_{j}||_{2}},
\end{aligned}
\end{equation}
where $\bm{s}_{k}$ is the semantic vector of the $k$-th class, $C$ is the number of classes, $\mathbf{S}^{s}$ and $\mathbf{S}^{c}$ are cosine similarity matrices of the class semantic vectors and center features, respectively.
We use GloVe~\cite{pennington2014glove} to generate semantic vectors of classes from their names.
The lower score implies that the trained feature space embeds semantic knowledge of the classes.

The evaluation results are in \Cref{SMCtable:feature_space}.
For the comparison, we train a network that follows our training strategy but using KCL for the contrastive loss.
Our method forms better latent space than the other. 
Furthermore, the feature space of SMC is more similar to the semantic information space.
SMC helps the network understand semantic information.

\begin{table}[t]
    \centering
    \begin{tabularx}{0.95\linewidth}{>{\hsize=.25\hsize}X | >{\hsize=.1\hsize}>{\centering}X  >{\hsize=.15\hsize}>{\centering}X  >{\hsize=.1\hsize}>{\centering}X | >{\hsize=.12\hsize}>{\centering\arraybackslash}X}         \hline
         Method &  Many & Medium & Few & All \\
         \hline
         \multicolumn{4}{l}{Inter-class score} \\
        Resize-KCL   & $0.578$ & $0.573$ & $0.596$ & $0.582$ \\
         Resize-SMC   & $\bm{0.573}$ & $\bm{0.564}$ & $\bm{0.590}$ & $\bm{0.575}$ \\
         \hline
         \hline
         \multicolumn{4}{l}{Semantic similarity score} \\
        Resize-KCL & $0.591$ & $0.581$ & $0.620$ & $0.596$ \\ 
         Resize-SMC   & $\bm{0.578}$ & $\bm{0.568}$ & $\bm{0.609}$ & $\bm{0.584}$ \\
         \hline
         \hline
         \multicolumn{4}{l}{Classification accuracy (\%)} \\
         Resize-KCL & $\bm{61.9}$ & $46.4$ & $31.9$ & $47.6$ \\
         Resize-TSC & $\bm{61.9}$ & $\bm{48.8}$ & $\bm{33.9}$ & $\bm{48.9}$ \\
         \hline
         
    \end{tabularx}
    \caption{The feature space analysis on CIFAR-100-LT (100). Lower scores and higher accuracy are better.}
    \label{SMCtable:feature_space}
\end{table}

\subsection{Component analysis}
In this subsection, we present an in-depth empirical analysis of our method to validate our proposals.
\Cref{SMCtable:ablation_cifar100} displays the comparison results, and the detailed description of the analysis follows.

\textbf{Normalized combination ratio in the data blending.}
We normalize the combination ratio to the range of $[ 0.2, 0.8 ]$ to preserve the foreground information and increase the data variance. We assess the effect of the restriction and show the result in the first block of \Cref{SMCtable:ablation_cifar100}. The constrained range help contain the information of original images, which enlarges the data variance. This observation emphasizes the impact of data diversity to the network.

\textbf{Information loss in cropping.}
The second block of \Cref{SMCtable:ablation_cifar100} is the comparison result between different data mixing methods. Crop-SMC replaces the resizing operation in SMC into random cropping operation.
Using resizing operation shows better performance than using random cropping, which implies that resizing is advantageous than random cropping for long-tailed recognition.

\textbf{Data augmentation analysis.}
We measure the importance of random cropping to create positive instances with alternative augmentations and present the comparison result in the third block of \Cref{SMCtable:ablation_cifar100}.
In line with the previous contrastive learning studies~\cite{chen2020simple}, the random cropping dramatically enhances the recognition performance. However, cropping on mixed images inhere the possibility of the information loss. Our augmentation method reduces the information loss, which results in the better performance.

\textbf{Adaptive weights on positives.}
In \cref{SMCeq:smc_loss}, we weight positives based on their combination ratios.
We assess the influence of the adaptive weights by replacing the weighted summation in \cref{SMCeq:smc_loss} with different loss weighting methods. The fourth block of \Cref{SMCtable:ablation_cifar100} illustrates the result.
Averaging indicates the simple averaging of the three losses, and the assigning uses the class of the larger component as the class label of the blended sample, then it trains the network with the na\"ive SCL.
If the combination ratio is high, the mixtures are more likely to have the foreground information than the background information.
The weights adjust the impact of a positive pair to the network to boost the performance of the model.
Furthermore, the lower performance of the assigning method emphasizes the necessity to define the positive and negative pair definitions for blended images.

\textbf{Loss hyperparameter analysis.}
We set the loss weight hyperparameter $\eta$ in \cref{SMCeq:total_loss} to $0.1$ in experiments. The last block of \Cref{SMCtable:ablation_cifar100} shows the performance dependence on the loss weight. Higher weight makes the contrastive objective dominate during training, while lower weight lessens the effect of the contrastive term. To fully utilize the power of SMC, the balance between the two losses is required.

\begin{table}[t]
    \centering
    \begin{tabularx}{0.95\linewidth}{X  S}
         \hline
         Comparisons & Acc (\%) \\
         \hline
         \multicolumn{2}{l}{Range of the combination ratio} \\
         $[ 0.0, 1.0 ]$ & $48.1$ \\
         $[ 0.2, 0.8 ]$ (proposed) & $48.9$ \\
         \hline
         \multicolumn{2}{l}{Information loss in cropping} \\
         Crop-SMC & $47.6$ \\
         Resizing-SMC (proposed) & $48.9$ \\
         \hline
         \multicolumn{2}{l}{Augmentation method for positives} \\
         No cropping & $47.3$ \\
         Cropping after mixing & $47.8$ \\
         Cropping before mixing (proposed) & $48.9$ \\
         \hline
         \multicolumn{2}{l}{Weights on positives} \\
         Averaging & $46.7$ \\
         Assigning the larger class label & $47.1$ \\
         Weighted summation (proposed) & $48.9$ \\
         \hline
         \multicolumn{2}{l}{Loss hyperparameter} \\
         $\eta=0.05$ & $47.0$\\
         $\eta=0.1$ & $48.9$ \\
         $\eta=0.2$ & $47.2$ \\
         \hline
    \end{tabularx}
    \caption{Component analysis of SMC on CIFAR-100-LT (100).}
    \label{SMCtable:ablation_cifar100}
\end{table}

\subsection{Ensembling on SMC}
The aforementioned comparisons have been compared on a single network. In this subsection, we show that SMC can benefit from ensemble-based methods.
We combine SMC with RIDE~\cite{RIDE}, a state-of-the-art ensemble-based long-tailed recognition method, and measure the performance differences followed by the change of the number of experts in the ensemble of networks.
We added an additional MLP head for each expert and trained the feature extractor with our training loss during the first stage training of RIDE.
\Cref{SMCtable:ensemble} shows the experimental result.  
As the number of experts increases, the performance gets better. This result demonstrates the expandability of SMC. Our method can be easily merged with additional long-tailed recognition methodologies for better recognition capability.

\begin{table}[t]
    \centering
    \begin{tabular}{l | c c c c }
        \hline
        Method & Many & Medium & Few & All\\
        \hline
        SMC (1 expert) & $61.9$ & $48.8$ & $33.9$ & $48.9$ \\
        2 experts      & $67.5$ & $50.9$ & $27.1$ & $49.6$ \\
        3 experts      & $70.5$ & $50.7$ & $25.8$ & $50.7$ \\
        4 experts      & $71.6$ & $50.6$ & $32.3$ & $52.5$ \\
        \hline
    \end{tabular}
    \caption{Classification accuracy (\%) comparison of SMC with the multi-expert ensemble method on CIFAR-100-LT (100).}
    \label{SMCtable:ensemble}
\end{table}

\begin{figure}[t]
    \centering
    \includegraphics[width=0.95\linewidth]{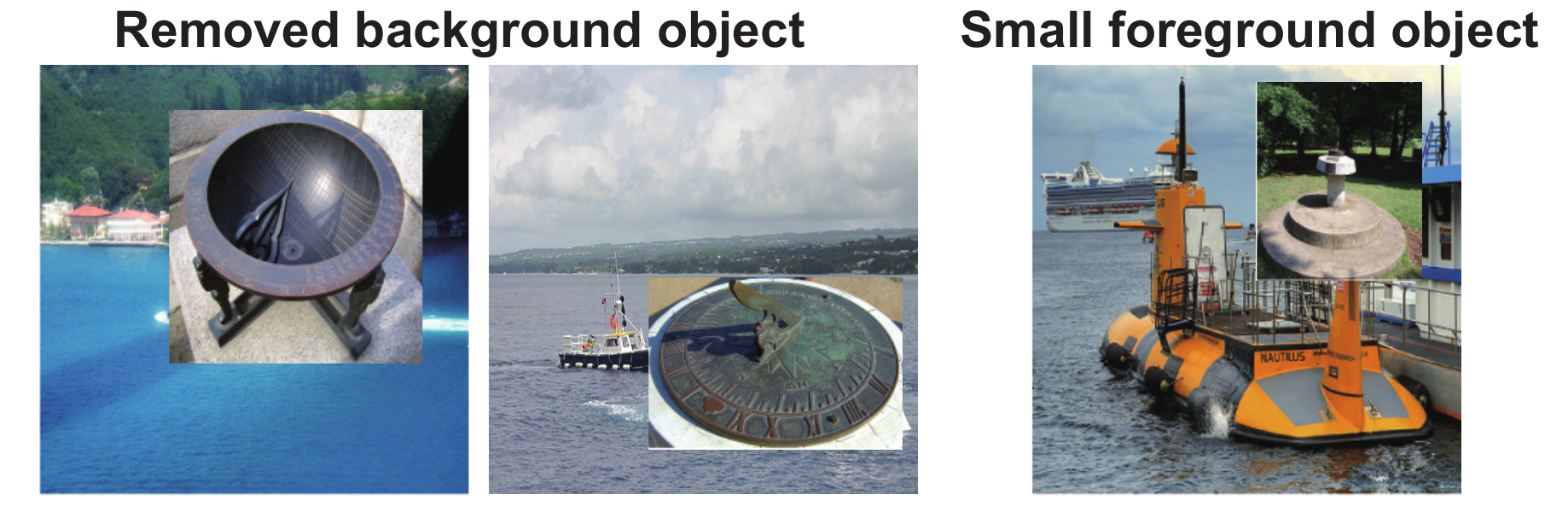}
    \caption{Visualization of information loss cases. The foreground and background classes are sundial and submarine, respectively. Although the restricted combination ratio and resizing operation pretend to preserve the image information, the information could be removed in the mixture. It is undesirable in the perspective of the data diversity. }
    \label{SMCfig:failures}
\end{figure}

\subsection{Limitations}
Our data augmentation method aims to preserve the semantic characteristics of training data.
Despite of the efforts to reduce the information losses in the training samples, the loss of critical information could occur.
\Cref{SMCfig:failures} illustrates the several failure cases.
The merging operation with a mask could remove the background object, which is undesirable in terms of the data diversity. Moreover, the resizing operation makes the object of interest in a foreground image too small that could drop the geometric information of the foreground object.
We believe that the performance of SMC could be improved with a proper knowledge-preserving augmentation method.

\section{Conclusion}
In this paper, we have introduced a novel long-tailed recognition method, named Supervised contrastive learning on Mixed Classes (SMC). Our method considers the information of blended images that enhances the variety of training data. The resizing operation on foregrounds helps the network explore more diverse data space. Furthermore, we define positive and negative pairs for mixed images and suggest a data augmentation method for the contrastive learning of blended images on long-tailed recognition. Performance comparisons on the various benchmarks and the in-depth analysis of our method validate our idea.

\textbf{Potential negative social impact}
The diversified training data requires longer training, and the additional supervised contrastive term requires larger memory. Thus, SMC may need more computational resources than the other methods, which could increase carbon emissions.

{\small
\bibliographystyle{ieee_fullname}
\bibliography{main_bib}
}

\end{document}